\documentclass{ieeeaccess}
\usepackage{graphicx} 

\usepackage{amsmath,amssymb,amsfonts,amsthm}
\usepackage{algorithmic}
\usepackage{textcomp}
\usepackage{booktabs,tabularx}
\usepackage{subfig}

\usepackage{float,graphicx,cite,enumerate}
\usepackage{listings}
\usepackage{hyperref}
\usepackage[table,xcdraw]{xcolor}
\usepackage{hyphenat}
\usepackage{bbold}
\usepackage[export]{adjustbox}
\usepackage{color,soul}

\usepackage{pmboxdraw}

\usepackage{draftwatermark}
\SetWatermarkText{ENHANCED VERSION}
\SetWatermarkScale{0.4}

\def\BibTeX{{\rm B\kern-.05em{\sc i\kern-.025em b}\kern-.08em
    T\kern-.1667em\lower.7ex\hbox{E}\kern-.125emX}}
    
\begin{document}
\history{Received 16 December 2025, accepted 9 February 2026, date of publication 18 February 2026. This work is an enhanced version of the
article accepted and published in IEEE
Access. Date of current version 4 March 2026}
\doi{Digital Object Identifier 10.1109/ACCESS.2026.3666073}

\title{How Does Fourier Analysis Network Work? A Mechanism Analysis and a New Dual-Activation Layer Proposal}

\author{Sam Jeong\authorrefmark{1}, and
Hae Yong Kim\authorrefmark{1}}
\address[1]{Dept. Electronic Systems Engineering, Polytechnic School, University of São Paulo, São Paulo, SP, 05508-010, Brazil.}
\tfootnote{This work was financed in part by the Coordenação de Aperfeiçoamento de Pessoal de Nível Superior (CAPES) Brasil [Finance Code 001, PhD scholarship 88887.838214/2023-00]; by the National Council for Scientific and Technological Development (CNPq) Brasil [grant 300724/2025-0]; and by São Paulo State Research Foundation (FAPESP) Brasil [process number 2024/10263-3].\\
The Article Processing Charge (APC) of this research publication was funded by the Coordenação de Aperfeiçoamento de Pessoal de Nível Superior – CAPES [ROR identifier: 00x0ma614].}


\corresp{Corresponding author: Sam Jeong (e-mail: sam.jeong@usp.br).}

\begin{abstract}
Fourier Analysis Network (FAN) was recently proposed as a simple way to improve neural network performance by replacing part of Rectified Linear Unit (ReLU) activations with sine and cosine functions. 
Although several studies have reported small but consistent gains across tasks, the underlying mechanism behind these improvements has remained unclear. 
In this work, we show that only sine contributes positively to performance, whereas cosine tends to be detrimental.
Our analysis reveals that the improvement is not a consequence of the sine function’s periodic nature; instead, it stems from the function’s local shape in the neighborhood of $x{=}0$.
We further show that sine activation primarily alleviates the dying-ReLU problem, in which a neuron consistently receives negative inputs, produces zero gradients, and stops learning. 
Although modern ReLU-like activations eliminate zero-gradient region, they still contain input domains where gradients remain diminished, hindering rapid convergence.
Sine activation addresses this limitation by introducing a more stable gradient pathway. This analysis leads to the development of the Dual-Activation Layer (DAL), a more efficient convergence accelerator. 
We evaluate DAL on three tasks: classification of noisy sinusoidal signals versus pure noise, MNIST digit classification, and Electrocardiogram (ECG)-based biometric recognition. 
In all instances, DAL-based models exhibit faster convergence rates relative to models using conventional activations.
Furthermore, DAL achieved slightly higher final test accuracy across all three tasks. 
This improvement was statistically significant for MNIST and ECG-ID; however, while a performance gain was observed in the simpler sinusoidal classification task, it did not reach the threshold for statistical significance.
\end{abstract}


\titlepgskip=-21pt

\maketitle

\section{Introduction}

Recently, Dong et al. \cite{Dong2025} introduced the Fourier Analysis Network (FAN). 
In essence, a FAN layer is a fully connected layer composed of 50\% ReLU-type activations (or similar variants such as Leaky ReLU, GELU, or Swish), 25\% sine activations, and 25\% cosine activations. 
The authors showed experimentally that replacing standard dense layers with FAN layers across a variety of tasks — including symbolic formula representation, time-series forecasting, language modeling, and image recognition — yields small yet consistent performance gains. 
As of December 2025, the paper has received 39 citations on Google Scholar, with several follow-up studies reporting similar performance improvements when employing FAN. 
Jeong et al. \cite{Jeong2025} extended this idea to convolutional architectures through the CFAN (Convolutional Fourier Analysis Network) and likewise observed modest gains across several problems.

Despite these empirical results, there has been no convincing explanation of why FAN improves performance. 
One intuitive hypothesis — that sine and cosine activations, being periodic functions, might better capture periodic structures in signals or images — does not hold up under scrutiny: the periodicity of interest in data occurs in time or space, whereas sine and cosine in FAN are periodic with respect to the activation input. 
Thus, there is no direct alignment between the periodicity of the activation and that of the data.

\begin{figure}[ht]
\centering
\begin{tabular}{ccc}
    \includegraphics[width=0.3\linewidth]{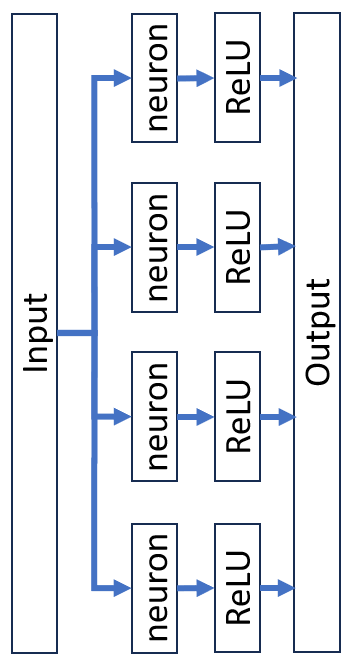} &
    \includegraphics[width=0.3\linewidth]{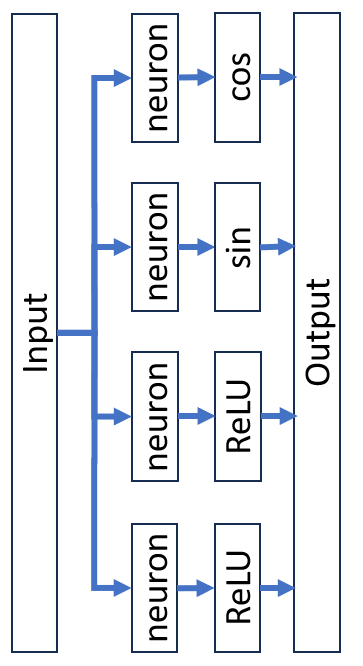} &
    \includegraphics[width=0.3\linewidth]{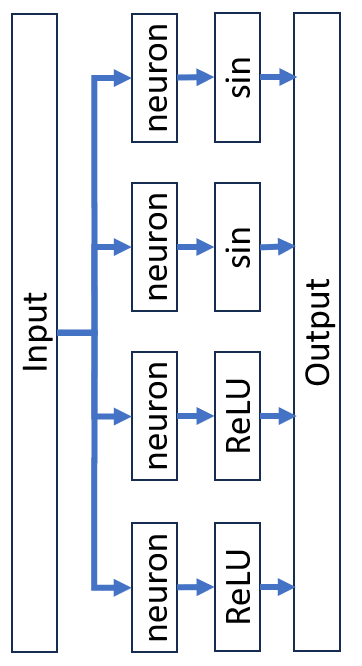} \\
    
    (a) & (b) & (c)
\end{tabular}
\caption{Schematic representation of activation layers composed of four neurons: (a) ReLU-only; (b) FAN (ReLU, sin, cos, 2:1:1); (c) DAL (ReLU, sin, 1:1).}
\label{fig:RELU_FAN_DAL}
\end{figure}

In this paper, we clarify the mechanism underlying FAN through simple 1D signal-classification experiments that distinguish between noisy sinusoids and pure noise. 
Our main findings are:

\begin{enumerate}
\item {\it Only the sine activation contributes positively.}
We show that the sine activation consistently improves network performance, whereas the cosine activation tends to degrade it. 
We further demonstrate that the improvement originates from the sine function's non-zero derivative at the origin rather than its periodic nature, providing a critical pathway for continuous optimization. 
Building on this insight, we propose the Dual-Activation Layer (DAL), denoted as $(f,g,a{:}b)$, which mixes two activation functions $f$ and $g$ in proportion $a{:}b$. 
Here, $f$ is ReLU (or a similar variant), and $g$ is sine (or another function with a similar shape near $x=0$, such as $tanh$).
Figure \ref{fig:RELU_FAN_DAL} provides a simplified comparison between the ReLU-only, FAN, and DAL activation layers for a single-channel configuration. 
In multi-channel layers, this structure is replicated on a per-channel basis. 
Each activation function is statically assigned to a specific neuron throughout the model's lifecycle, from training through to inference.

\item \textit{DAL accelerates convergence in a statistically significant manner.}
For the problem studied, the configuration (ReLU, Sine, 6:2) significantly outperforms the ReLU-only baseline during the early training epochs.
Similar gains are observed when ReLU and sine are replaced by other activation functions with comparable local shapes.
We emphasize, however, that the optimal ratio $a{:}b$ is task- and architecture-dependent.

\item {\it DAL mitigates the dead-neuron problem.}
We show that DAL enhances performance by counteracting the “dead” (or “nearly dead”) neuron phenomenon associated with activation $f$, thereby ensuring a more stable gradient flow during training.
\end{enumerate}

Finally, we apply DAL to real-world tasks — handwritten digit classification (MNIST) and electrocardiogram-based biometric recognition (ECG-ID) — and observe consistent acceleration in convergence and statistically significant higher final accuracy across both problems.

\section{Related works}

The choice of activation functions plays a central role in the performance, stability, and convergence speed of neural networks. 
Early architectures relied almost exclusively on sigmoids and hyperbolic tangent functions, but the introduction of the Rectified Linear Unit (ReLU) by Nair and Hinton \cite{Nair2010ReLU} established a new default due to its simplicity and favorable gradient properties. 
However, ReLU suffers from the well-known “dying neuron” problem, in which neurons receiving consistently negative inputs produce zero output and zero gradient, permanently halting learning. 
Several variants have been proposed to alleviate this issue, such as Leaky ReLU \cite{Maas2013LeakyReLU}, Parametric ReLU (PReLU) \cite{He2015PReLU}, Exponential Linear Units (ELU) \cite{Clevert2016ELU}, Gaussian Error Linear Units (GELU) \cite{Hendrycks2016GELU}, and Swish \cite{Ramachandran2017Swish}. 
These functions soften the hard zero-gradient region of ReLU and have been shown to improve convergence in a variety of settings.

Beyond monotonic activations, several works have explored the use of periodic or oscillatory functions in neural networks. Periodic activations have been investigated in the context of implicit neural representations
where sinusoidal activations are essential for modeling complex signals, as in SIREN \cite{Sitzmann2020SIREN}. 
Other works have used Fourier features or encodings to enrich input representations and improve generalization \cite{Tancik2020FourierFeatures}. 
However, these approaches are fundamentally distinct from FAN. 
They primarily employ sinusoidal transformations on inputs or features for enhanced representational capacity or complex function modeling, whereas FAN's mechanism relies on mixing sinusoidal and ReLU-like functions as activations to directly improve the network's training dynamics and overall performance.

FAN architecture proposed by Dong et al. \cite{Dong2025} introduced a hybrid activation layer composed of ReLU, sine, and cosine functions. FAN demonstrated small but consistent improvements across diverse tasks, including language modeling, time-series forecasting, and symbolic expression learning. 
Despite these empirical gains, the underlying mechanism behind the improved performance remained unclear. 
Dong et al. hypothesized that the periodic nature of sine and cosine could help capture periodic patterns in data, though no evidence supporting this explanation was provided.

Subsequent studies have examined the use of FAN across a range of application domains.
Dere et al. \cite{Dere2025} incorporated FAN layers into the fully connected stages of an electromyography gesture-recognition network (EMG-Net), reporting a reduction in gesture-classification error from 12.22\% to 6.92\%.
Zeng et al. \cite{Zeng2025} replaced the fully connected layers in a passive seismic source-localization framework with FAN layers; however, because their work does not include a clear comparison against baseline architectures, the magnitude of the improvement remains difficult to assess.

Other studies have explored variants and extensions of FAN within transformer architectures and Kolmogorov-Arnold Networks (KAN) \cite{Tkhai2025, Eslamian2025, Wang2025, Tan2025}.
Wang et al. \cite{Wang2025} report that incorporating FAN into a transformer increases the mean accuracy for  electroencephalogram (EEG)-based emotion recognition from 67.78\% to 70.30\%.
However, the large and highly overlapping standard deviations (9.69 vs. 9.89) raise doubts about the statistical significance of this improvement.
Similarly, Tan et al. \cite{Tan2025} show that a Fourier-based KAN improves the average Dice similarity coefficient on a hepatic vessel segmentation task from 66.57\% to 67.57\%, yet no statistical analysis is provided to assess whether this difference is meaningful.
This uncertainty is further reinforced by the findings of Eslamian et al. \cite{Eslamian2025}, who evaluated multiple KAN variants across diverse tasks and concluded that FAN-based KANs do not consistently achieve state-of-the-art performance. 
Their results suggest that the gains reported in some studies may fall within natural performance variability rather than reflecting a systematic or robust advantage.

Jeong et al. \cite{Jeong2025} proposed the Convolutional FAN (CFAN), extending FAN to convolutional layers, and likewise observed only modest gains.

Although several works attempt to implement FAN or similar ideas, few provide evidence of statistically significant improvements or a clear explanation of the mechanism by which FAN would offer an advantage. 

In parallel, several lines of research have focused on improving gradient propagation in deep networks. 
Residual connections \cite{He2016ResNet}, normalization layers \cite{Ioffe2015BatchNorm}, and carefully designed activation functions have all been shown to help mitigate gradient attenuation and dead-neuron phenomena. 
Works analyzing the flow of gradients through activation functions argue that the local shape of the activation — especially near zero — strongly influences trainability \cite{glorot2010understanding, dubey2022activation}.

Despite these advances, to the best of our knowledge, no prior work has investigated the hypothesis that the empirical improvements reported for FAN arise not from periodicity, but from the local properties of the sine function near zero, nor has any work evaluated cosine and sine independently within the FAN structure. 
Likewise, the proposed Dual-Activation Layer (DAL), which explicitly combines a ReLU-like activation with a smooth zero-centered function (such as sine or tanh), appears to be novel.

Thus, this work contributes to the literature by:

\begin{enumerate}
\item Providing the first systematic study separating the roles of sine and cosine in FAN;
\item Demonstrating that improvements stem from improved gradient flow and reduced dead neurons, not from periodic representational capacity; and
\item Introducing DAL, a principled and general mechanism for mixing activation functions to improve convergence across diverse architectures and tasks.
\end{enumerate}

\section{Methodology}

\subsection{Problem Setup}

To investigate the effect of different activation functions, we consider an extremely simple 1D signal classification task implemented using Keras/TensorFlow. 
We generate two types of 64-sample signals:

\begin{itemize}
\item {\it Sine:} Signals consisting of six periods of a sinusoid with random phase and amplitude between 0.5 and 1.0, to which Gaussian noise $N(0,1)$ is added at every sample.
\item {\it Noise:} Purely noisy signals, where each sample is drawn independently from $N(0,1)$.
\end{itemize}

\begin{figure}
    \centering
    \includegraphics[width=1\columnwidth]{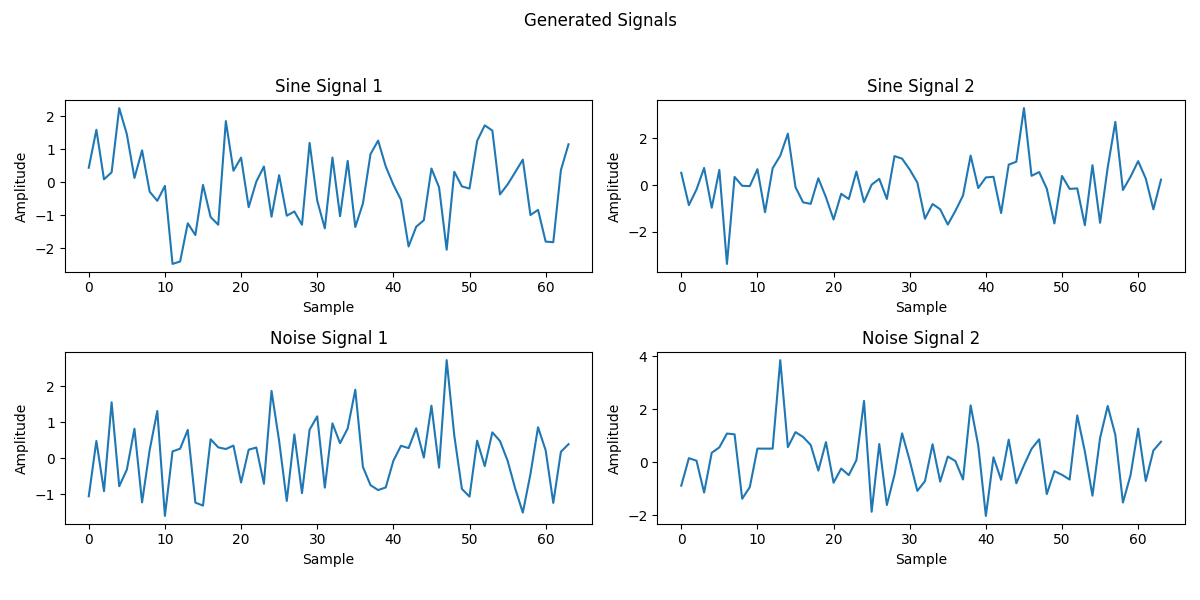}
    \caption{Examples of signals used in the experiments.}
    \label{nv01_graph}
\end{figure}

Figure \ref{nv01_graph} illustrates two examples of each signal type (nv01.py).
In our experiments, we compare two network architectures, CNNA and CNNB, on the task of classifying each signal as sine or noise. 
For each trial, we generate 100 signals from each class, shuffle them, and split the dataset in half: 50 for training and 50 for testing. 
Both networks are trained and evaluated on exactly the same set of signals. 
Training lasts for $M$ epochs, and the entire procedure is repeated 
$N$ times, yielding $N$ paired test accuracies. 
We then apply one-sided paired t-test \footnote{scipy.stats.ttest\_rel} to assess whether the differences between CNNA and CNNB are statistically significant.

Additionally, we compute the mean performance curves — accuracy and loss for both training and testing — averaged over the $N$ repetitions across the $M$ epochs. 
These curves allow us to visualize the learning dynamics and compare the convergence speed of the two architectures.

\subsection{Network Architecture}

We employ a simple and widely used 1D CNN architecture, illustrated in below. 

{\scriptsize 
\begin{verbatim}
│ Layer                 │ Output Shape    │ Param #  │
│ InputLayer            │ (None, 64, 1)   │       0  │
│ Conv1D (kernel=5)     │ (None, 64, 16)  │      96  │
│ MaxPooling1D (size=2) │ (None, 32, 16)  │       0  │
│ Conv1D (kernel=5)     │ (None, 32, 32)  │   2,592  │
│ MaxPooling1D (size=2) │ (None, 16, 32)  │       0  │
│ Conv1D (kernel=5)     │ (None, 16, 64)  │  10,304  │
│ MaxPooling1D (size=2) │ (None, 8, 64)   │       0  │
│ Conv1D (kernel=5)     │ (None, 8, 64)   │  20,544  │
│ MaxPooling1D (size=2) │ (None, 4, 64)   │       0  │
│ Conv1D (kernel=5)     │ (None, 4, 64)   │  20,544  │
│ MaxPooling1D (size=2) │ (None, 2, 64)   │       0  │
│ Flatten               │ (None, 128)     │       0  │
│ Dense                 │ (None, 64)      │   8,256  │
│ Dense (sigmoid)       │ (None, 1)       │      65  │
\end{verbatim}
} 

The architecture comprises five 1D convolutional blocks (kernel size = 5, “same” padding), each featuring an activation layer followed by a max-pooling layer with a pool size of 2.
After the convolutional stack, two dense layers are applied. 
The output layer always uses a sigmoid activation, while the activation functions of the five convolutional layers and the first dense layer vary depending on the experiment (FAN, DAL or single activation).

To ensure a fair comparison, CNNA and CNNB are identical in every respect — including initialization with the same pseudo-random weights — except for the activation functions used. Training is performed using the Adam optimizer and the binary crossentropy loss function.

This architecture was chosen purely as a controlled environment to isolate and study the effect of activation functions. 
In these initial experiments, we intentionally omitted Batch Normalization (BN) layers. 
BN is known to stabilize training and mitigate the “dying ReLU” problem. 
By excluding it, we created a controlled environment to isolate the specific effect of the activation functions on gradient stability, which is the central mechanism analyzed in this work.
Higher accuracies could certainly be obtained with more sophisticated architectures (incorporating regularization, residual connection, data augmentation, refined hyperparameter tuning, etc.).

\subsection{Reproducibility}

The effects observed in this study are subtle and sensitive to small variations. 
Readers attempting to reproduce the experiments should be aware that the results may differ not only when applied to other problems, but even when replicating the exact same task, depending on the programming environment, random initialization, and software versions used. 

All programs associated with this article are available at a shared folder \footnote{\scriptsize \url{https://drive.google.com/drive/folders/1ScArmAPIaOvxzbrEeIi-4HFmWnjstaNY}}, and their filenames are referenced throughout the text.

\section{Experiments and Results}

\subsection{FAN(ReLU, sin, cos, 6:1:1) vs. ReLU-only, $M{=}6$}

We begin our experiments by evaluating the original idea behind FAN: combining a conventional activation function (such as ReLU) with sine and cosine activations. 
To this end, we compare a FAN architecture that uses ReLU, sine, and cosine in a 6:1:1 ratio against a model that uses ReLU exclusively. 
We adopted the 6:1:1 proportion — rather than the originally suggested 2:1:1 — because it yielded slightly better results in our preliminary tests.
We performed $N{=}100$ independent training runs, each with $M{=}6$ epochs, obtaining the following results (nv17.py):

{\scriptsize 
\begin{verbatim}
Mean Test Accuracy (CNNA - relu/sin/cos 6:1:1): 0.7638
Mean Test Accuracy (CNNB - relu): 0.7378
One-sided (>) paired p-value: 0.0150
\end{verbatim}
}

These results indicate, with statistical significance $(p{=}0.0150)$, that replacing a fraction of the ReLU activations with sine and cosine leads to improved performance. 
However, a key question remains:
Is the improvement driven primarily by the sine activation, by the cosine activation, or by the combination of both?
The experiments in the following sections address this question directly.

\subsection{(ReLU, cos, 6:2) vs. ReLU-only, $M{=}6$}

We now evaluate the effect of using only the cosine activation, completely removing the sine activation. To this end, we compare (ReLU, cos, 6:2) with the baseline ReLU-only. 
Running $N{=}50$ trials, each with $M{=}6$ training epochs, we obtained (nv19.py):

{\scriptsize 
\begin{verbatim}
Mean Test Accuracy (CNNA - relu/cos 6:2): 0.6804
Mean Test Accuracy (CNNB - relu): 0.7600
One-sided (<) paired p-value: 0.0000
\end{verbatim}
}

Substituting a portion of ReLU activations with cosine functions leads to a significant degradation in performance ($p{<}0.001$). 
This result suggests that the cosine activation negatively impacts network efficacy in this configuration.

\subsection{DAL(ReLU, sin, 6:2) vs. ReLU-only, $M{=}6$}
Next, we evaluate the effect of combining ReLU with the sine activation, this time excluding cosine entirely. 
We compare DAL (ReLU, sin, 6:2) against the baseline ReLU-only. 
Running $N{=}100$ trials, each with $M{=}6$ training epochs, we obtain (nv21.py):

{\scriptsize 
\begin{verbatim}
Mean Test Accuracy (CNNA - relu/sin 6:2): 0.8090
Mean Test Accuracy (CNNB - relu): 0.7338
One-sided (>) paired p-value: 0.0000
\end{verbatim}
}

\begin{figure}
    \centering
    \includegraphics[width=1\columnwidth]{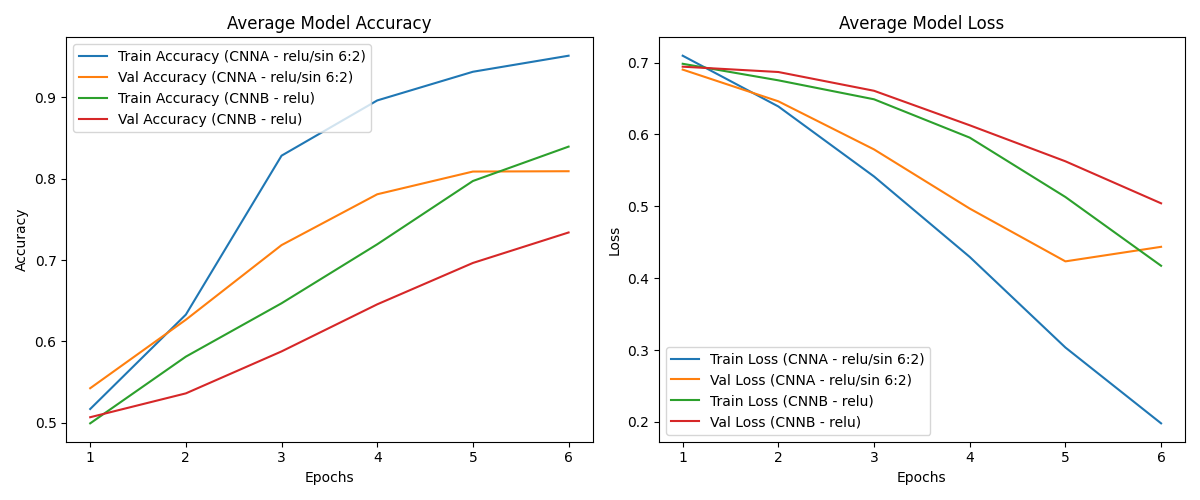}
    \caption{Evolution of performance metrics (averaged over $N{=}100$ runs) for DAL(ReLU, sin, 6:2) and ReLU-only. 
    The former exhibits consistently superior performance.}
    \label{nv21_graph}
\end{figure}

This highly significant improvement ($p{<}0.001$) shows that the sine activation is the primary contributor to the performance gains previously attributed to FAN.
Figure \ref{nv21_graph} illustrates the average evolution of the performance metrics over the 6 training epochs, aggregated across all 100 runs. The model using DAL converges more quickly and consistently outperforms the pure-ReLU model throughout training.

While Keras and TensorFlow utilize Glorot (Xavier) Uniform as the default initialization, we extended our experiments to include He (Kaiming) Normal and Uniform distributions. In all cases, the DAL architecture demonstrated superior convergence rates compared to the ReLU-only baseline, regardless of the initialization policy employed.

\subsection{DAL(ReLU, sin, 6:2) vs. ReLU-only, $M{=}60$} 

\begin{figure}
    \centering
    \includegraphics[width=1\columnwidth]{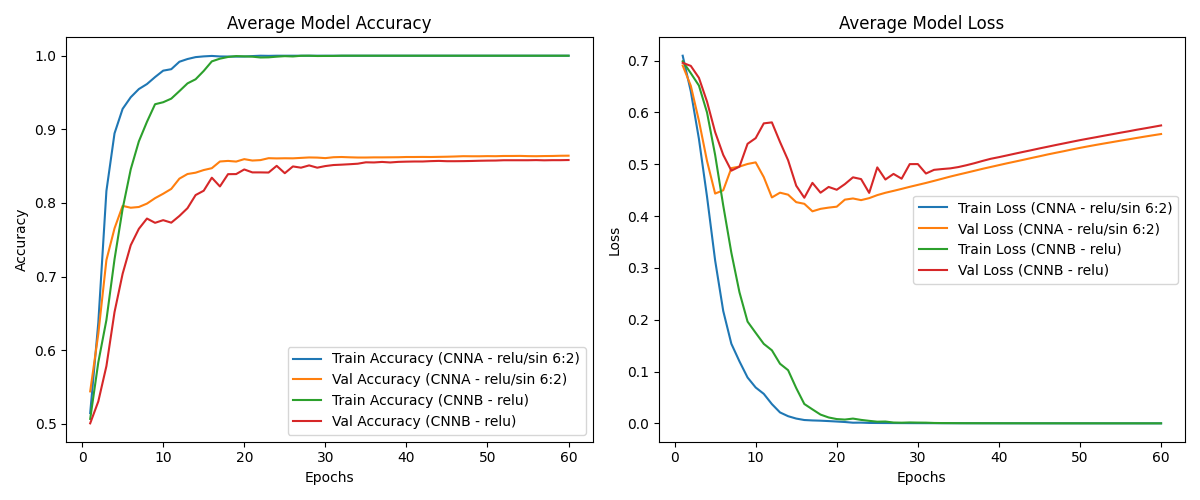}
    \caption{Evolution of performance metrics averaged over $N{=}100$ runs across $M{=}60$ epochs for DAL(ReLU, sin, 6:2) and ReLU-only.}
    \label{nv21f_graph}
\end{figure}

The advantage of DAL over pure-ReLU diminishes when the training is extended.
Repeating the same experiment $N{=}100$ times, now with $M{=}60$ epochs, we obtain (Figure \ref{nv21f_graph}, nv21f.py):

{\scriptsize 
\begin{verbatim}
Mean Test Accuracy (CNNA - relu/sin 6:2): 0.8642
Mean Test Accuracy (CNNB - relu): 0.8583
One-sided (>) paired p-value: 0.1137
\end{verbatim}
}

The final accuracy achieved by the DAL model is higher than that of the ReLU-only baseline. 
However, the resulting $p$-value of 0.1137 is insufficient to conclude that the DAL significantly outperforms the baseline in terms of final performance. 
On the other hand, the DAL architecture exhibits a significantly higher convergence rate, as illustrated in Figure \ref{nv21f_graph}. 
Such an acceleration is of considerable practical value and should not be overlooked.
The 6:2 ratio was not optimized, and different proportions may yield better results. 

\section{Discussion}

\subsection{Non-periodic Activations}

\begin{figure}
    \centering
    \includegraphics[width=1\columnwidth]{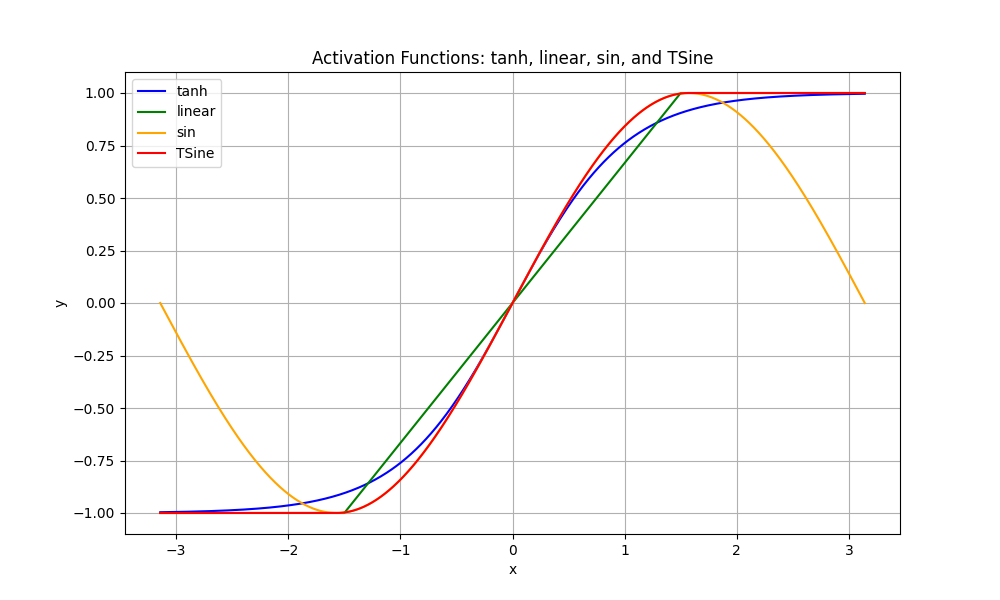}
    \caption{Activation functions tanh, linear, sin, and TSine.}
    \label{plot_activations}
\end{figure}

Why does replacing part of the ReLU activations with sine accelerate convergence? 
One hypothesis is that, because sine is periodic, it might better represent a sinusoidal signal contaminated by noise. 
To test whether periodicity is truly responsible for the improvement, we replaced the sine function with three non-periodic functions: tanh, truncated sine (TSine), and linear. 
The TSine and linear functions are defined below, and their plots are shown in Figure \ref{plot_activations}.

\[
\mathrm{TSine}(x) =
\begin{cases}
-1, & x \le -\frac{\pi}{2}, \\[6pt]
\sin(x), & -\frac{\pi}{2} < x \le \frac{\pi}{2}, \\[6pt]
1, & x > \frac{\pi}{2}.
\end{cases}
\]

\[
\mathrm{linear}(x) =
\begin{cases}
-1, & x \le -1.5, \\[6pt]
\frac{2}{3}x, & -1.5 < x \le 1.5, \\[6pt]
1, & x > 1.5.
\end{cases}
\]

\begin{figure}
    \centering
    \includegraphics[width=1\columnwidth]{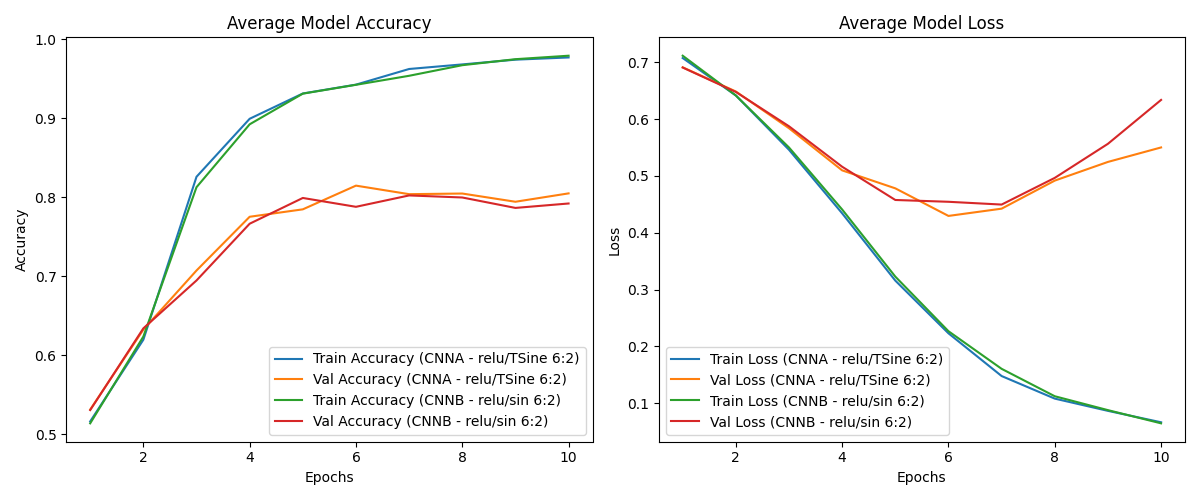}
    \caption{Comparison of (ReLU, TSine, 6:2) vs. (ReLU, sin, 6:2) activation configurations.}
    \label{nv23_graphs}
\end{figure}

\begin{figure}
    \centering
    \includegraphics[width=1\columnwidth]{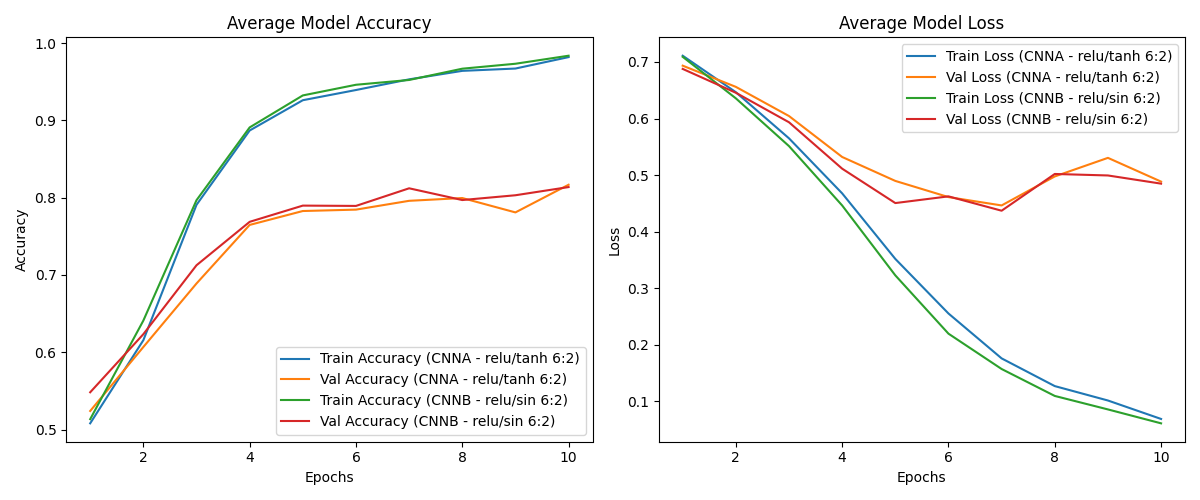}
    \caption{Comparison of (ReLU, tanh, 6:2) vs. (ReLU, sin, 6:2).}
    \label{nvb01_graphs}
\end{figure}

\begin{figure}
    \centering
    \includegraphics[width=1\columnwidth]{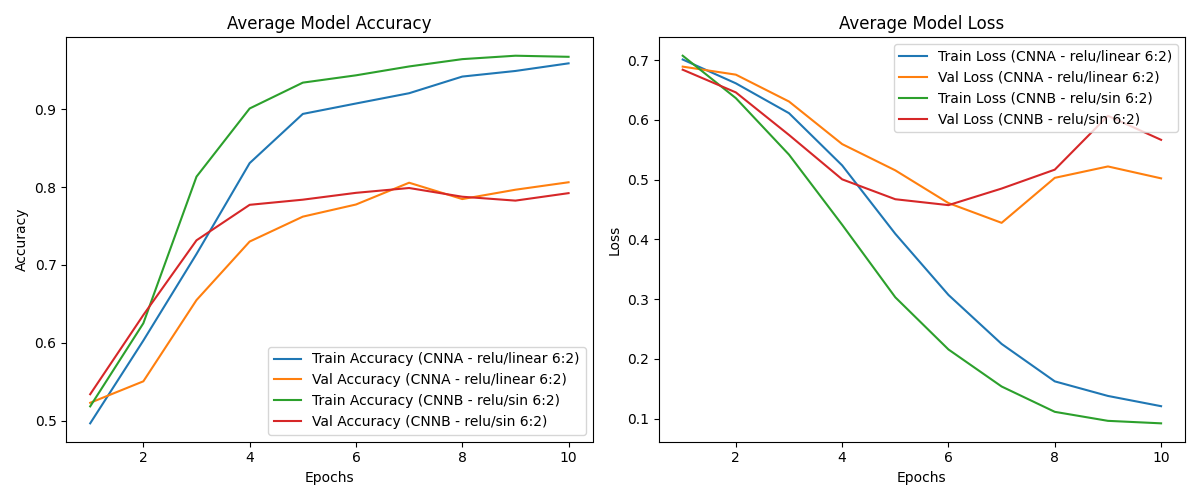}
    \caption{(ReLU, linear, 6:2) vs. (ReLU, sin, 6:2)}
    \label{nvb02_graphs}
\end{figure}

We evaluated (ReLU, $f$, 6:2) for $f {\in}$ \{TSine, tanh, linear\}, and compared the results with (ReLU, sin, 6:2). 
Running $N{=}50$ tests with $M{=}10$ epochs each, we obtained the results shown in Figures \ref{nv23_graphs}, \ref{nvb01_graphs}, and \ref{nvb02_graphs} (nv23.py, nvb01.py, nvb02.py). 
Networks using TSine and tanh achieved performance levels virtually identical to those obtained with sine, whereas the linear activation exhibited slightly slower convergence.

Since the improvements persist even when sine is replaced by non-periodic functions that nonetheless resemble sine within the approximate interval $x {\in} [ -1.5, +1.5 ]$, we conclude that the benefit does not arise from periodicity. 
Instead, it comes from the local shape of the function around $x{=}0$.

Most neural network frameworks are designed to perform optimally when the input data are normalized. 
Consequently, most practical implementations include a pre-processing step that normalizes the inputs.
In addition, popular weight initialization methods are designed to preserve an approximately standard normal distribution across the internal layers. 
Under this assumption, most values entering the activation functions are expected to lie roughly within the interval $x \in [-1.5, +1.5]$, which highlights the importance of the activation function’s shape within this range.
By empirically computing the statistics of the pre-activation values in the program used in this section, we obtained $\mu \approx 0$ for all layers, $\sigma \approx 0.45$ for the convolutional layers, and $\sigma \approx 0.85$ for the dense layer (nv90.py).
These results indicate that, in practice, the standard deviation is smaller than one, further reinforcing the importance of the activation function’s behavior in the interval $x \in [-1.5, +1.5]$.

\subsection{Dead Neurons}

Why does replacing some of the ReLU activations with sine, TSine, or tanh speed up training?
Because these alternative activations help mitigate the well-known dead neuron problem. 
A ReLU neuron is considered “dead” when its input remains consistently negative for all training samples. 
Since ReLU outputs zero for non-positive inputs, a dead neuron always produces zero output, and its derivative is also zero. 
With a zero derivative, that neuron’s weights and bias stop being updated during backpropagation.
Moreover, a dead neuron blocks gradient flow: the gradient passed backward to the previous layer is multiplied by zero, preventing the neurons feeding into it from learning.

To test this hypothesis, we trained the models DAL (ReLU, sin, 6:2) and ReLU-only, each $N{=}20$ times for $M{=}20$ epochs (nv81.py). 
We then computed and printed, for each trained network, the average percentage of dead neurons per layer across the 20 runs.

Table \ref{dead_neurons} shows that DAL (ReLU, sin, 6:2) exhibits substantially fewer dead neurons than ReLU alone. 
The alternative activation (sine) provides an additional gradient pathway, reducing the number of dead neurons. 
Incorporating sine yields a more robust gradient flow and lowers the likelihood of neuron death during training.
Consequently, given two architectures with identical neuron counts, the DAL model maintains a higher population of active neurons in practice than the ReLU-only model. 
Utilizing DAL is therefore functionally equivalent to increasing model capacity without expanding the actual parameter count.

An implication of this reasoning is that TSine should be avoided in practice: it contains wide regions with zero gradient, which can also lead to dead neurons, much like ReLU.
\vspace{0.2cm}

\begin{table}[h]
\centering
\begin{tabular}{lcc}
\hline
\textbf{Layer} & \textbf{(ReLU, sin, 6:2)} & \textbf{Only ReLU} \\
\hline
conv1  & 0.00\%  & 0.00\%  \\
conv2  & 0.00\%  & 0.00\%  \\
conv3  & 1.02\%  & 2.81\%  \\
conv4  & 2.42\%  & 7.81\%  \\
conv5  & 4.61\%  & 15.00\% \\
dense1 & 16.80\% & 36.95\% \\
\hline
\end{tabular}
\caption{Percentage of dead neurons per layer for the models (ReLU, sin, 6:2) and only ReLU.}
\label{dead_neurons}
\end{table}

\subsection{Modern Activation Functions}

\begin{figure}
    \centering
    \includegraphics[width=1\columnwidth]{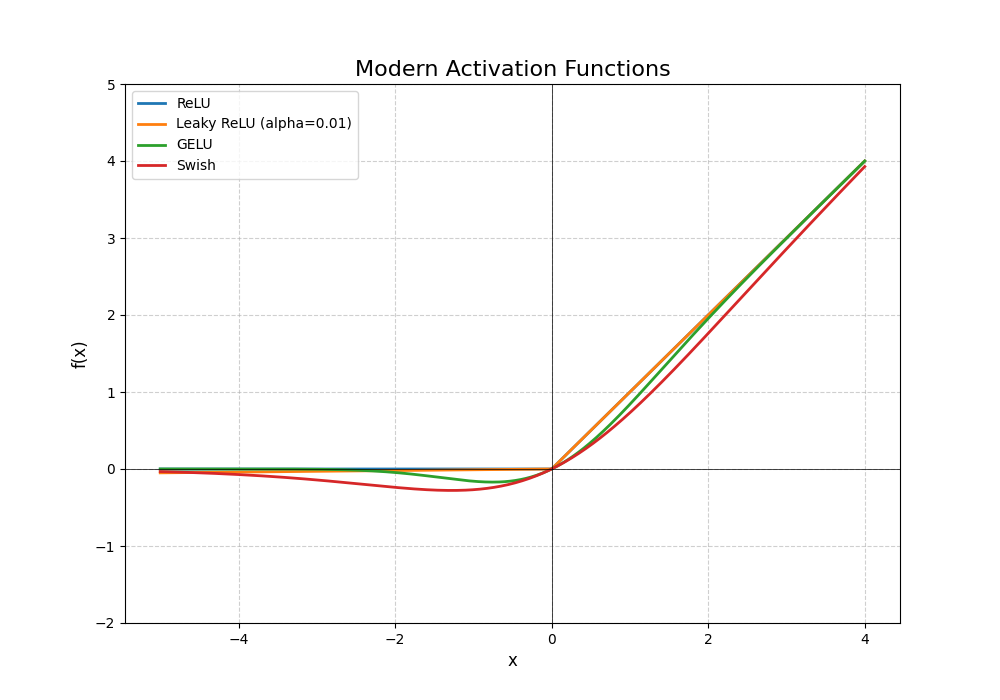}
    \caption{Modern activation functions.}
    \label{plot_moderna}
\end{figure}

We have seen that replacing a portion of the ReLU activations with sine or tanh accelerates network training by reducing the proportion of “dead” neurons and improving gradient flow. 
However, several modern activation functions have already been proposed in the literature to mitigate this issue, including Leaky ReLU \cite{Maas2013LeakyReLU}, GELU \cite{Hendrycks2016GELU}, and Swish \cite{Ramachandran2017Swish} (Figure \ref{plot_moderna}). 
These modern activations do not contain regions with zero derivative. 
The question, then, is whether DAL can also accelerate convergence when these modern activations are used.
Even though these activation functions avoid dead neurons entirely, their shapes remain similar to ReLU, meaning they still contain regions with small (though non-zero) derivatives. 
Thus, it is plausible that replacing a subset of these activations with sine might accelerate convergence.

\begin{figure}
    \centering
    \includegraphics[width=1\columnwidth]{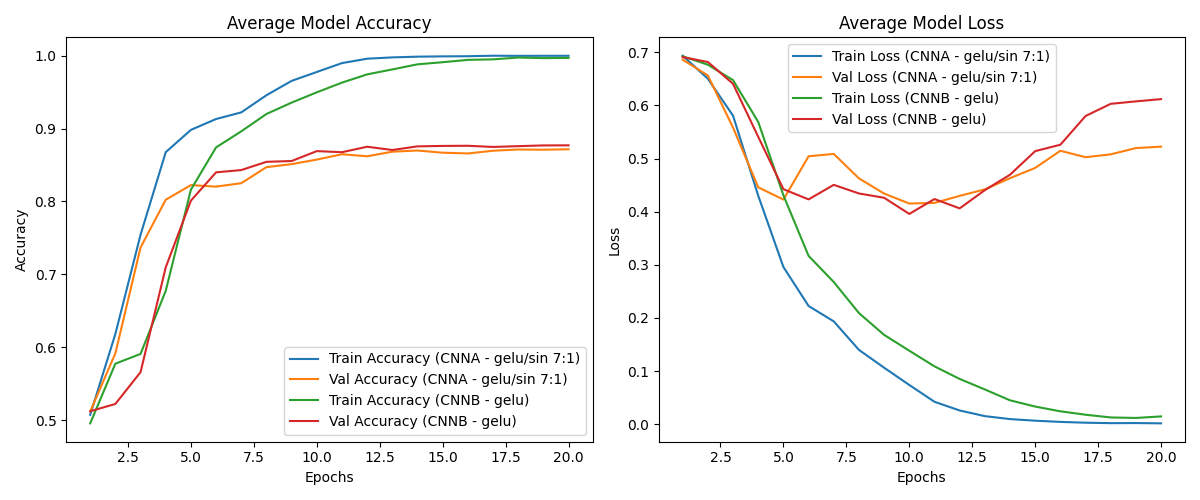}
    \caption{(GELU, sin, 7:1) converges faster than GELU-only.}
    \label{nv60c_graphs}
\end{figure}

\begin{figure}
    \centering
    \includegraphics[width=1\columnwidth]{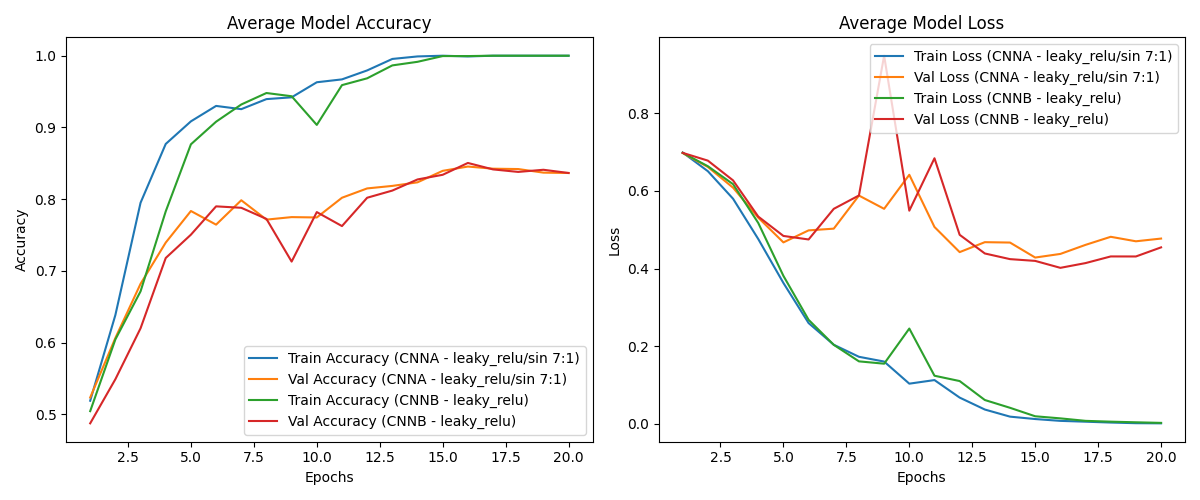}
    \caption{(Leaky Relu, sin, 7:1) converges faster than Leaky Relu-only.}
    \label{nv60f_graphs}
\end{figure}

\begin{figure}
    \centering
    \includegraphics[width=1\columnwidth]{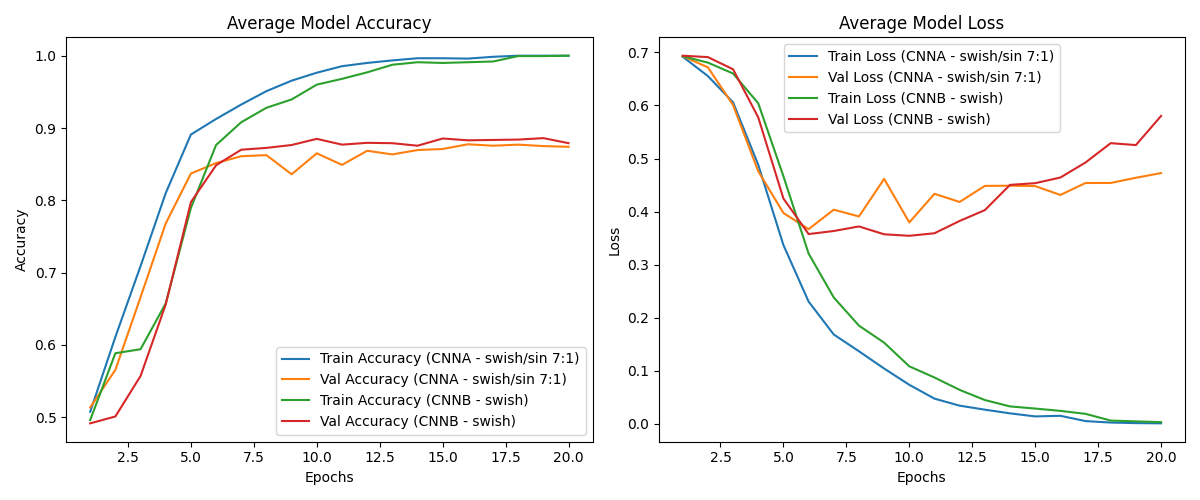}
    \caption{(Swish, sin, 7:1) converges faster than Swish-only.}
    \label{nv60e_graphs}
\end{figure}

To test this hypothesis, we compare (GELU, sin, 7:1) against GELU-only (nv60c.py). 
Figure \ref{nv60c_graphs} shows the results over $M{=}20$ epochs, averaged across $N{=}100$ runs. 
Although both models reach similar final accuracies, (GELU, sin, 7:1) converges noticeably faster than the pure GELU model. 
We observed similar behavior with Leaky ReLU (Figure \ref{nv60f_graphs}, nv60f.py) and Swish (Figure \ref{nv60e_graphs}, nv60e.py).

\section{Real-World Applications}

\subsection{MNIST Classification}
\label{MNIST_Classification}

To assess whether the conclusions drawn from the synthetic 1D classification problem generalize to a more realistic task, we evaluated handwritten digit classification on the MNIST dataset using convolutional neural networks with activation DAL (ReLU, sin, 6:2) and pure ReLU.
Both models share the following architecture:

{\scriptsize 
\begin{verbatim}
│ Layer             │ Output Shape        │  Param # │
│ Normalization     │ (None, 28, 28, 1)   │        3 │
│ Conv2D 5×5        │ (None, 24, 24, 20)  │      520 │
│ Activation        │ (None, 24, 24, 20)  │        0 │
│ MaxPooling2D 2×2  │ (None, 12, 12, 20)  │        0 │
│ Conv2D 5×5        │ (None, 8, 8, 40)    │   20,040 │
│ Activation        │ (None, 8, 8, 40)    │        0 │
│ MaxPooling2D 2×2  │ (None, 4, 4, 40)    │        0 │
│ Flatten           │ (None, 640)         │        0 │
│ Dense             │ (None, 200)         │  128,200 │
│ Activation        │ (None, 200)         │        0 │
│ Dense (softmax)   │ (None, 10)          │    2,010 │
\end{verbatim}
}

We implemented a Keras/TensorFlow program that performs the following steps:

\begin{enumerate}
\item Creates two models, CNNA and CNNB, with the architecture shown above, for training and evaluating MNIST classification.
\item Ensures that the two models are identical in all aspects — including initialization with the same pseudo-random weights — except for the activation functions.
\item Except for the final softmax layer, all activation layers in CNNA use a combination of ReLU and sine functions in a 6:2 ratio, whereas CNNB uses ReLU exclusively.
\item Uses a final dense layer with 10 neurons and softmax activation to perform digit classification.
\item Includes a preprocessing (normalization) layer in both models, which standardizes the input images using the mean and standard deviation computed from the training set.
\item Trains both models using the Adam optimizer and categorical cross-entropy loss.
\item After each training epoch, computes the training and test accuracies and losses for both models.
\item Plots the evolution of all four metrics (training accuracy, test accuracy, training loss, and test loss) for both networks.
\end{enumerate}

\subsubsection{MNIST classification for $M{=}5$ epochs}

\begin{figure}
    \centering
    \includegraphics[width=1\columnwidth]{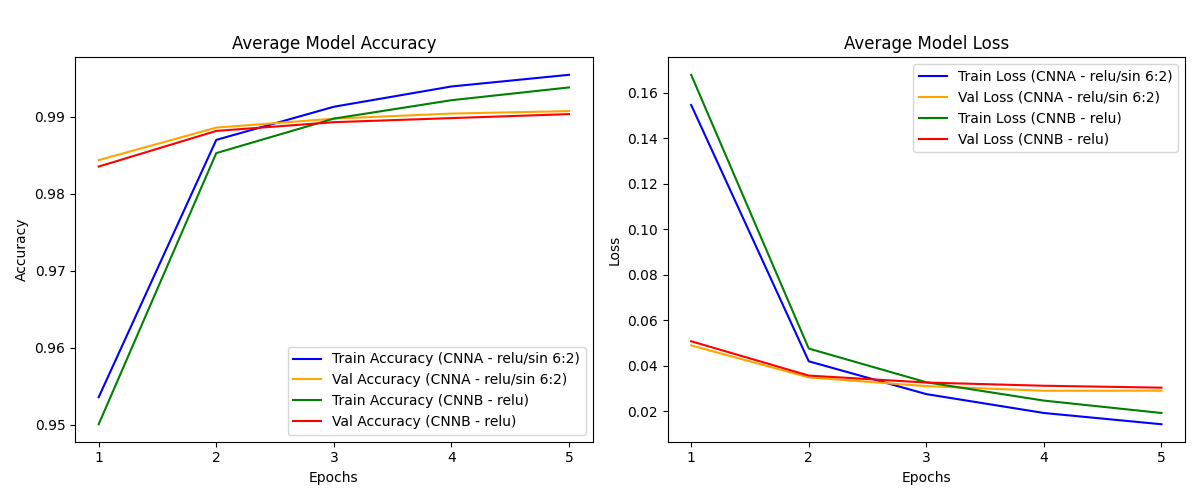}
    \caption{Performance metrics during the first four training epochs of the MNIST classifier.}
    \label{da10_metrics_comparison}
\end{figure}

Given the near-ceiling performance levels on this task, a ceiling effect obscures the performance gap, making it difficult to determine whether one algorithm statistically outperforms the other.
Nevertheless, training both models for only $M{=}5$ epochs over $N{=}50$ runs reveals statistically significant disparities: the differences in training and test accuracy yielded $p$-values of $p{<}0.001$ and $p{=}0.0280$, respectively.
This indicates that DAL indeed converges faster than ReLU-only (Figure \ref{da10_metrics_comparison}, da10.py).

{\scriptsize 
\begin{verbatim}
Training Accuracy Comparison:
CNNA (ReLU, Sine): 0.9957
CNNB (ReLU): 0.9939
One-sided (>) paired p-value: 0.0000
\end{verbatim}
}

{\scriptsize 
\begin{verbatim}
Test Accuracy Comparison:
CNNA (ReLU, Sine): 0.9910
CNNB (ReLU): 0.9905
One-sided (>) paired p-value: 0.0280
\end{verbatim}
}

\subsubsection{MNIST classification for $M{=}30$ epochs}

Even when training both models for $M{=}30$ epochs over $N{=}50$ repetitions, the disparity in final training and test accuracy persists and remains statistically significant ($p{=}0.0020$ and $p{=}0.0165$).

{\scriptsize 
\begin{verbatim}
Training Accuracy Comparison:
CNNA (ReLU, Sine): 0.9994
CNNB (ReLU): 0.9991
One-sided (>) paired p-value: 0.0020
\end{verbatim}
}

{\scriptsize 
\begin{verbatim}
Test Accuracy Comparison:
CNNA (ReLU, Sine): 0.9923
CNNB (ReLU): 0.9918
One-sided (>) paired p-value: 0.0165
\end{verbatim}
}

\begin{figure}
    \centering
    \includegraphics[width=1\columnwidth]{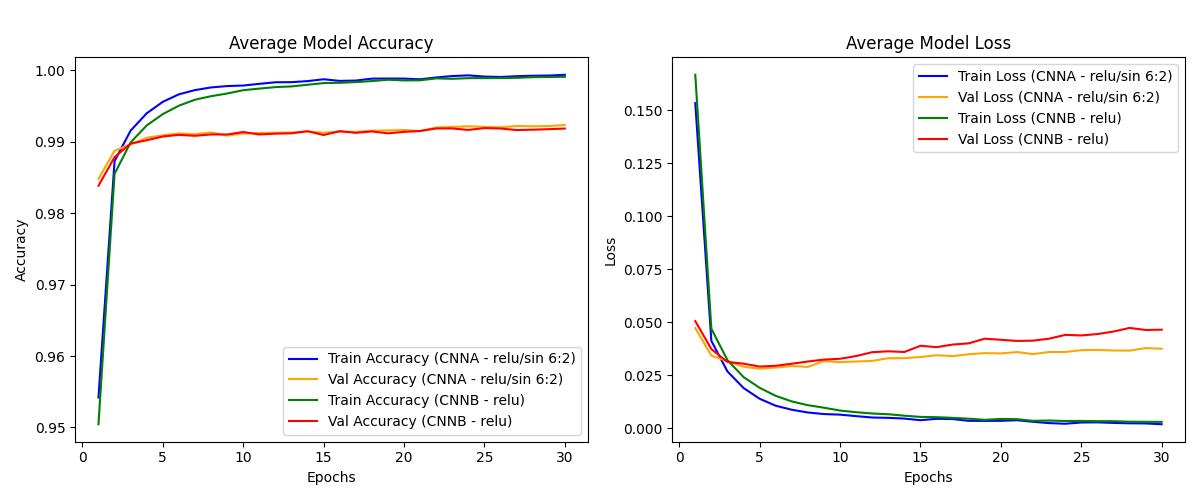}
    \caption{Training and test accuracies and losses over $M{=}30$ epochs for MNIST classification, averaged over $N{=}50$ runs. 
    Comparison between (ReLU, sin, 6:2) and ReLU-only.}
    \label{da11_metrics_comparison}
\end{figure}

When we examine the test losses (Figure \ref{da11_metrics_comparison}), both models reach a minimum and then begin to worsen, most likely due to overfitting.
However, overfitting appears to be less severe in the DAL model than in the pure ReLU model.

\subsection{ECG-ID Classification}
\label{ECG_ID}

We also evaluated a more complex 1D signal classification task: identity-recognition using the ECG-ID dataset \cite{ecgid}, available through the PhysioNet repository \cite{Physionet}.
The ECG-ID dataset contains 310 single-lead ECG recordings from 90 individuals, each 20 seconds long and sampled at 500 Hz.
R-peaks were detected using the Pan–Tompkins algorithm \cite{Pan1985}. Following the segmentation procedure described by Nemirko and Lugovaya \cite{ecgid}, each cardiac cycle was extracted by taking an 80-sample window before the R-peak and a 170-sample window after it.

For each recording, the eight cardiac cycles with the smallest Euclidean distance to the subject-specific average cycle were selected.
Each selected cycle was then mean-centered and divided by standard deviation.
This process yielded a total of 2,456 cardiac cycles across 90 classes (one class per individual).

The dataset was partitioned into two stratified folds. For each run, one fold was used for training and the remaining fold was used to evaluate.

\subsubsection{ECG-ID Classification Without Batch Normalization}

We implemented a PyTorch program that follows the same experimental procedure used for MNIST classification (Section \ref{MNIST_Classification}), using the following architecture:

{\scriptsize 
\begin{verbatim}
│ Layer                     │ Output Shape    │ Param #  │
│ InputLayer                │ (None, 1, 251)  │       0  │
│ Conv1D (k=25)             │ (None, 36, 251) │     936  │
│ AvgPool (size=5, strid=4) │ (None, 36, 63)  │       0  │
│ Conv1D (k=25)             │ (None, 36, 63)  │  32,436  │
│ [[BatchNormalization]]    │                 │          │
│ AvgPool (size=5, strid=4) │ (None, 36, 16)  │       0  │
│ Conv1D (k=25)             │ (None, 576, 16) │ 518,976  │
│ AdaptiveAvgPool1D         │ (None, 576, 1)  │       0  │
│ Dense                     │ (None, 120)     │  69,240  │
│ Dense                     │ (None, 84)      │  10,164  │
│ Dense (softmax)           │ (None, 90)      │   7,650  │
\end{verbatim}
} 

Excluding the final softmax layer, CNNA utilizes a 1:1 ratio of ReLU and Sine activations in all convolutional and dense layers, while CNNB is composed entirely of standard ReLU activations.

\begin{figure}
    \centering
    \includegraphics[width=1\columnwidth]{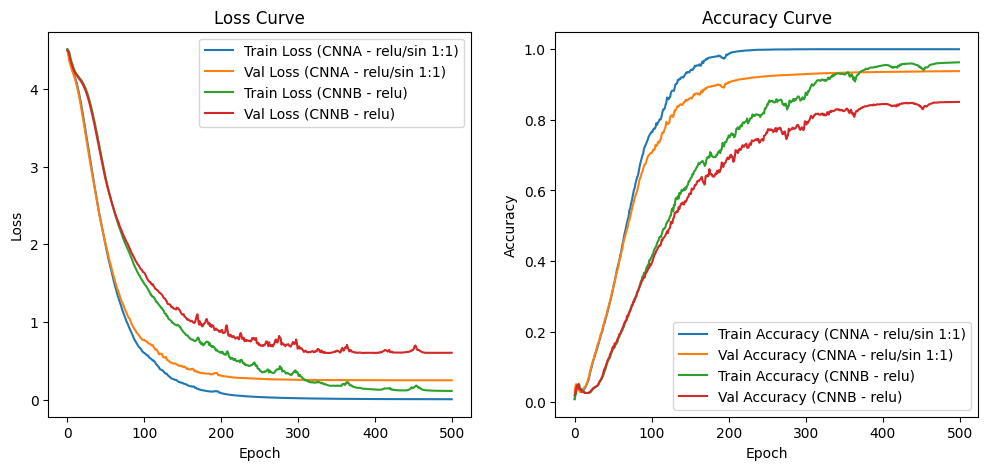}
    \caption{ 
    Evolution of training and test metrics for ECG-ID classification. 
    Results are averaged over $N{=}10$ independent runs for $M{=}500$ epochs without batch normalization. 
    The DAL architecture (ReLU, sin, 1:1) exhibits accelerated convergence and achieves statistically significant performance gains compared to the pure ReLU baseline.}
    \label{ECG_ID_graphs}
\end{figure}

Training both models for $M{=}500$ epochs across $N{=}10$ runs reveals statistically significant differences in both training and test accuracy for both short and long training regimes ($M{=}50$ and $M{=}500$).
Note that given the 90-class nature of the dataset, the baseline for random chance is approximately 1.1\%, placing the observed accuracy in its proper context.
These results show that, in this more complex problem, the DAL model converges faster and achieves higher final training and test accuracy compared to pure ReLU (Figure \ref{ECG_ID_graphs}, ecgid.ipynb).

{\scriptsize 
\begin{verbatim}
Training Accuracy Comparison with 50 epochs:
CNNA (ReLU, Sine): 0.3146 (Std Dev: 0.0264)
CNNB (ReLU): 0.1459 (Std Dev: 0.0244)
One-sided (>) paired p-value: 0.0000
The difference is statistically significant.

Test Accuracy Comparison with 50 epochs:
CNNA (ReLU, Sine): 0.3107 (Std Dev: 0.0268)
CNNB (ReLU): 0.1535 (Std Dev: 0.0246)
One-sided (>) paired p-value: 0.0000
The difference is statistically significant.

Training Accuracy Comparison with 500 epoch:
CNNA (ReLU, Sine): 1.0000 (Std Dev: 0.0000)
CNNB (ReLU): 0.9629 (Std Dev: 0.0125)
One-sided (>) paired p-value: 0.0000
The difference is statistically significant.

Test Accuracy Comparison with 500 epoch:
CNNA (ReLU, Sine): 0.9379 (Std Dev: 0.0093)
CNNB (ReLU): 0.8507 (Std Dev: 0.0206)
One-sided (>) paired p-value: 0.0000
The difference is statistically significant.
\end{verbatim}
}

The experiments in this section were implemented in PyTorch using the library's default weight initialization. 
To ensure robustness, we also evaluated the models by explicitly enforcing both Glorot (Xavier) uniform and He (Kaiming) uniform initializations across all layers. 
In all test scenarios — for both $M{=}50$ and $M{=}500$ epochs — the DAL model achieved statistically significant higher accuracy than the ReLU-only baseline.

\subsubsection{ECG-ID Classification With Batch Normalization}

\begin{figure}
    \centering
    \includegraphics[width=1\columnwidth]{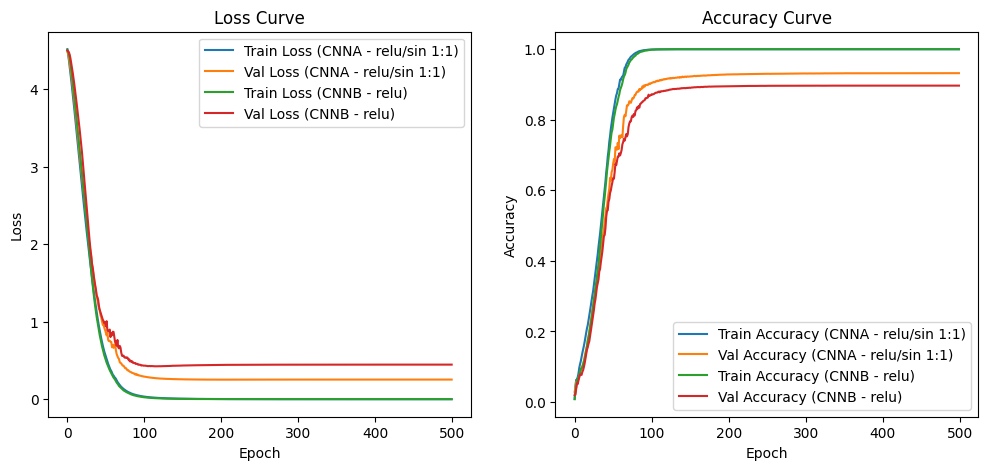}
    \caption{Comparison between (ReLU, sin, 1:1) and ReLU-only models, both including a batch normalization after the second convolutional block.
    Training was conducted for $M{=}500$ epochs and repeated over $N{=}10$ independent runs.
    }
    \label{ECG_ID_normalized}
\end{figure}

Integrating a batch normalization layer after the second convolutional block and prior to average pooling accelerates the convergence of both networks (compare Figures \ref{ECG_ID_graphs} and \ref{ECG_ID_normalized}). 
Under this configuration, the training accuracy curves for both models become nearly identical, suggesting that batch normalization compensates for much of
ReLU’s inherent instability. 
Regarding test performance, DAL consistently outperforms ReLU-only; while this lead is not statistically significant in the early stages, it achieves statistical significance once full convergence is attained.

{\scriptsize 
\begin{verbatim}

Training Accuracy Comparison with 50 epochs:
CNNA (ReLU, Sine): 0.8058 (Std Dev: 0.0397)
CNNB (ReLU): 0.7721 (Std Dev: 0.0569)
One-sided (>) paired p-value: 0.0125
The difference is statistically significant.

Test Accuracy Comparison with 50 epochs:
CNNA (ReLU, Sine): 0.6581 (Std Dev: 0.0506)
CNNB (ReLU): 0.6229 (Std Dev: 0.0896)
One-sided (>) paired p-value: 0.0768
The difference is not statistically significant.

Training Accuracy Comparison with 500 epochs:
CNNA (ReLU, Sine): 1.0000 (Std Dev: 0.0000)
CNNB (ReLU): 1.0000 (Std Dev: 0.0000)
p-value cannot be computed.

Test Accuracy Comparison with 500 epochs:
CNNA (ReLU, Sine): 0.9321 (Std Dev: 0.0145)
CNNB (ReLU): 0.8968 (Std Dev: 0.0208)
One-sided (>) paired p-value: 0.0000
The difference is statistically significant.

\end{verbatim}
}

\subsubsection{Dead neurons without and with batch normalization}

Table \ref{dead_neurons_ECGID_1} reports the percentage of dead neurons for models without batch normalization, whereas Table \ref{dead_neurons_ECGID_2} shows the corresponding values with batch normalization.
The results indicate that networks using DAL(ReLU, sin, 1:1) are notably less prone to neuron inactivation.
Without normalization, the ReLU-only model exhibits severe neuron death — reaching up to 75\% in dense layers — while the dual-activation model remains considerably more stable.

Batch normalization substantially reduces neuron death in both architectures.
Even so, the DAL model consistently presents lower dead-neuron rates, higher test accuracy  and lower test loss,
supporting the idea that combining ReLU with sine activations improves gradient flow compared to using ReLU alone.

Consequently, we conclude that, especially in deep and complex network architectures, the DAL can be effectively employed alongside other gradient-enhancing techniques, such as batch normalization and residual connections, to yield a synergistic effect.

\begin{table}[h]
\centering
\begin{tabular}{lcc}
\hline
\textbf{Layer} & \textbf{(ReLU, sin, 1:1)} & \textbf{Only ReLU} \\
\hline
conv1  & 0.00\%  & 0.00\%  \\
conv2  & 0.00\%  & 16.67\%  \\
conv3  & 8.16\%  & 33.86\%  \\
dense1  & 23.34\%  & 75.00\%  \\
dense2  & 7.15\%  & 41.67\% \\

\hline
\end{tabular}
\caption{Percentage of dead neurons per layer without batch normalization.}
\label{dead_neurons_ECGID_1}
\end{table}

\begin{table}[h]
\centering
\begin{tabular}{lcc}
\hline
\textbf{Layer} & \textbf{(ReLU, sin, 1:1)} & \textbf{Only ReLU} \\
\hline
conv1  & 0.00\%  & 0.00\%  \\
conv2  & 0.00\%  & 0.00\%  \\
BatchNormalization  & & \\
conv3  & 0.00\%  & 0.00\%  \\
dense1  & 15.00\%  & 52.50\%  \\
dense2  & 3.57\%  & 11.91\% \\

\hline
\end{tabular}
\caption{Percentage of dead neurons per layer with batch normalization.}
\label{dead_neurons_ECGID_2}
\end{table}

\section{Conclusions}

In this work, we investigated the mechanism underlying the performance improvements reported for FAN. 
Although FAN was originally interpreted as benefiting from the periodic nature of sine and cosine activations, our experiments demonstrate that this explanation is insufficient. 
By isolating the contribution of each activation function, we showed that only the sine activation contributes positively to performance, while the cosine activation consistently degrades convergence and accuracy.

Our results further reveal that the benefit of sine does not arise from periodicity, but rather from its local non-zero derivative shape near $x{=}0$. 
Activation functions such as tanh and truncated sine — despite being non-periodic — produce similar improvements when mixed with ReLU. 

A key finding of this work is that mixing ReLU with a smooth, zero-centered activation such as sine helps mitigate the dying ReLU problem, reducing the number of inactive neurons and enabling more stable gradient propagation.
This improved gradient flow leads to faster convergence, even in architectures that do not suffer from strictly zero gradients, such as those using GELU, Swish, or Leaky ReLU.
Based on these insights, we proposed the DAL, which combines two activation functions in a single layer — for example, ReLU and sine. 

When comparing two architectures of equal size, the DAL model demonstrates a higher density of active units than its conventional counterparts. 
Consequently, adopting the DAL approach enhances the network's effective width, providing the benefits of a larger architecture without the computational cost of additional parameters.

Across a diverse range of tasks — including synthetic 1D signal classification, MNIST digit recognition, and ECG-ID biometric recognition — the DAL consistently accelerates convergence. 
It frequently yields a final performance superior to standard activations, with these improvements reaching statistical significance in the majority of cases. 
Furthermore, the computational overhead introduced by the sine function is marginal, ensuring that the DAL maintains nearly the same efficiency as a conventional ReLU layer.

In summary, we have established that the performance gain attributed to FAN is not due to the spectral decomposition hypothesis, but rather to the introduction of activation functions that stabilize the gradient flow near the origin, specifically mitigating the “dying ReLU” problem. 
Our proposed DAL serves as a robust convergence accelerator. 

In very deep or complex network architectures, the DAL can be effectively employed in conjunction with other gradient flow enhancement techniques — such as batch normalization, skip connections, and advanced optimizers — to provide a synergistic effect. 
Consequently, the DAL represents a promising and straightforward architectural component to ensure more stable and rapid convergence, especially when combined with these established practices for building state-of-the-art models.

\bibliographystyle{IEEEtran}
\bibliography{refs}

\begin{IEEEbiography}[{\includegraphics[width=1in,height=1.25in,clip,keepaspectratio]{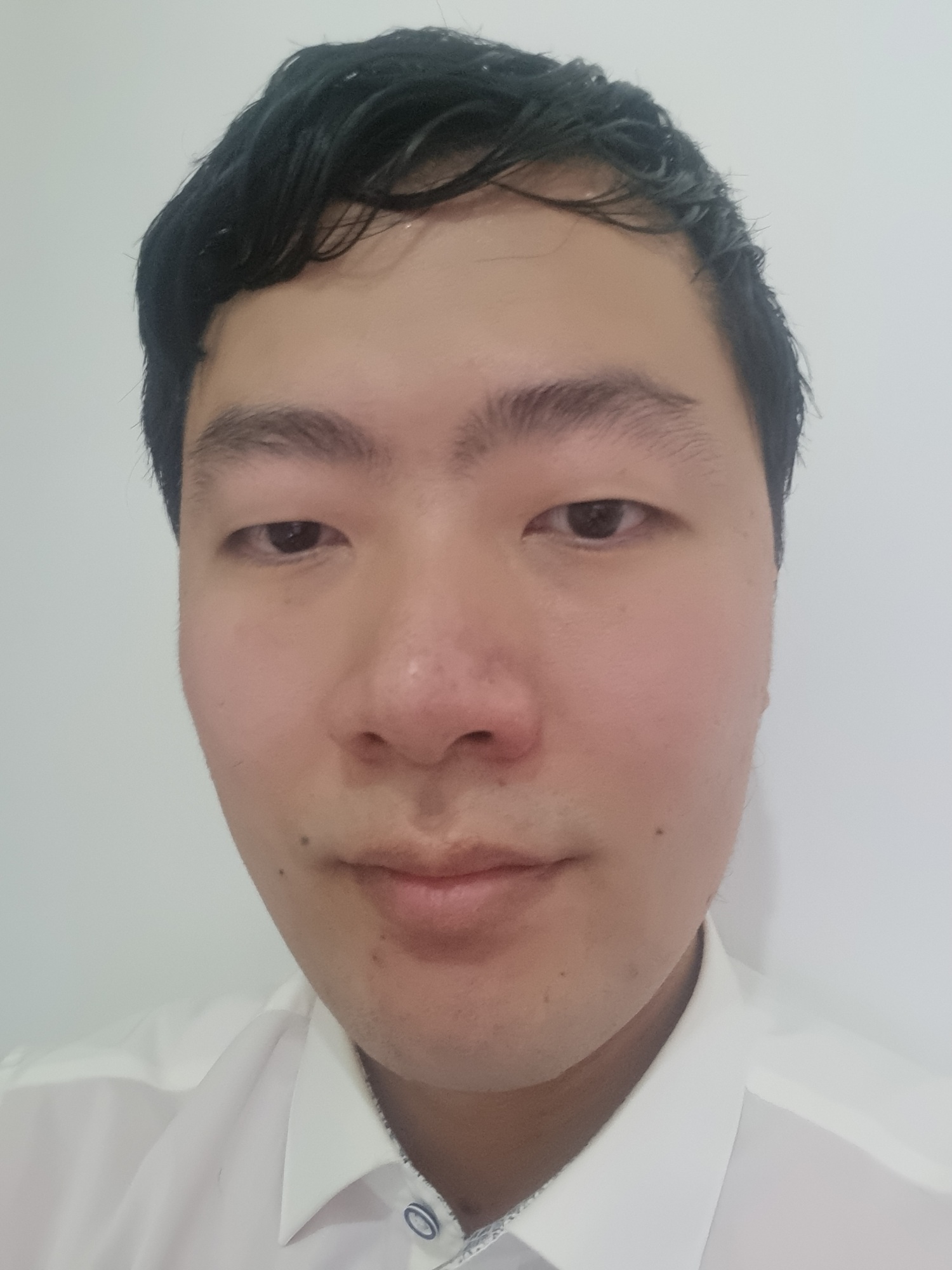}}]{Sam Jeong} received the B.S. degree in electrical engineering with emphasis in electronic systems in 2017, followed by M.S. degree in electrical engineering with area of concentration in biomedical engineering in 2021, both from the University of São Paulo (USP), Brazil. He is currently pursuing the Ph.D. degree in electrical engineering with the Department of Electronic Systems, USP.

His research interest includes signal and image processing, medical signal and image analysis, and machine learning.
\end{IEEEbiography}

\begin{IEEEbiography}[{\includegraphics[width=1in,height=1.25in,clip,keepaspectratio]{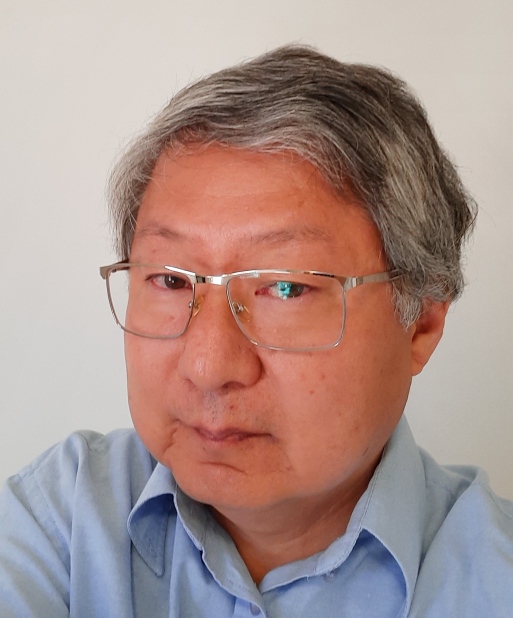}}]{Hae Yong Kim}
received the B.S. and M.S. degrees (with distinctions) in computer science and the Ph.D. degree in electrical engineering from the Universidade de São Paulo (USP), Brazil, in 1988, 1992 and 1997, respectively. 

He is currently an Associate Professor with the Department of Electronic Systems Engineering, USP. He is the author of more than 100 articles and holds three patents. His research interests include image processing, machine learning, medical image processing and computer security. 

Dr. Kim and colleagues received the 6th edition of the Petrobras Technology Award in the ``Refining and Petrochemical Technology'' category (2013); the ``Best Paper in Image Analysis'' award at the Pacific-Rim Symposium on Image and Video Technology (2007); and the Thomson ISI Essential Science Indicators ``Hot Paper'' award, for writing one of the top 0.1\% of the most cited computer science papers (2005).
\end{IEEEbiography}

\EOD

\end{document}